\documentclass[10pt,twocolumn,letterpaper]{article}

\usepackage{cvpr}
\usepackage{times}
\usepackage{epsfig}
\usepackage{graphicx}
\usepackage{amsmath}
\usepackage{amssymb}
\usepackage[mathscr]{euscript}

\usepackage{booktabs} %
\usepackage[small]{caption}

\usepackage{float}
\usepackage{aliascnt}
\usepackage{dsfont}
\usepackage{balance}

\usepackage[pagebackref=true,breaklinks=true,letterpaper=true,colorlinks,bookmarks=false]{hyperref}

\cvprfinalcopy %

\newtheorem{theorem}{Theorem}
\newtheorem{corollary}{Corollary}[theorem]

\newcommand{\ignore}[1]{}
\newcommand{\mbf}[1]{\mathbf{#1}}

\DeclareRobustCommand\code[1]{%
  \ifmmode
    \expandafter\texttt
  \else
    \expandafter\textnhtt
  \fi{#1}%
}

\newaliascnt{eqfloat}{equation}
\newfloat{eqfloat}{h}{eqflts}
\floatname{eqfloat}{Equation}

\newcommand*{\ORGeqfloat}{}
\let\ORGeqfloat\eqfloat
\def\eqfloat{%
  \let\ORIGINALcaption\caption
  \def\caption{%
    \addtocounter{equation}{-1}%
    \ORIGINALcaption
  }%
  \ORGeqfloat
}

\def\cvprPaperID{1831} %

\ifcvprfinal\fi
\begin{document}

\title{High-dimensional Convolutional Networks for Geometric Pattern Recognition}

\author{Christopher Choy\\
NVIDIA
\and
Junha Lee\\
POSTECH
\and
Ren\'{e} Ranftl\\
Intel Labs
\and
Jaesik Park\\
POSTECH
\and
Vladlen Koltun\\
Intel Labs
}

\maketitle

\begin{abstract}
Many problems in science and engineering can be formulated in terms of geometric patterns in high-dimensional spaces. We present high-dimensional convolutional networks (ConvNets) for pattern recognition problems that arise in the context of geometric registration. We first study the effectiveness of convolutional networks in detecting linear subspaces in high-dimensional spaces with up to 32 dimensions: much higher dimensionality than prior applications of ConvNets. We then apply high-dimensional ConvNets to 3D registration under rigid motions and image correspondence estimation. Experiments indicate that our high-dimensional ConvNets outperform prior approaches that relied on deep networks based on global pooling operators.
\end{abstract}

\vspace{-4mm}
\section{Introduction}

Finding structure in noisy data is a general problem that arises in many different disciplines. For example, robust linear regression requires finding a pattern (line, plane) in noisy data. 3D registration of point clouds requires the identification of veridical correspondences in the presence of spurious ones~\cite{Choi2015}. Structure from motion (SfM) pipelines use verification based on prescribed geometric models to filter spurious image matches~\cite{Schoenberger2016sfm}. A variety of such applications can benefit from improved methods for the detection of geometric structures in noisy data.

Such detection is challenging. Data points that belong to the sought-after structure often constitute only a small fraction, while the majority are outliers. Various algorithms have been proposed over the years to cope with noisy data~\cite{Brachmann2016,Chum2005,Fitzgibbon2003,Hoseinnezhad2011,Huber2011,Raguram2013,Rousseeuw1984,Tennakoon2016}, but they are usually specific to a subset of problems.

Recent works have advocated for using deep networks~\cite{Ranftl2018,Yi2018,Zhang2019} to learn robust models to classify geometric structures in the presence of outliers. Deep networks offer significant flexibility and the promise to replace hand-crafted algorithms and heuristics by models learned directly from data.
However, due to the unstructured nature of the data in geometric problems, existing works have treated such data as unordered sets, and relied on network architectures based predominantly on global pooling operators and multi-layer perceptrons (MLPs)~\cite{Qi2017,Zaheer2017deep}. Such network architectures lack the capacity to model local geometric structures and do not leverage the nature of the data, which is often embedded in a (high-dimensional) metric space and has a meaningful geometric structure.

In this work, we introduce a novel type of deep convolutional network that can operate in high dimensions. Our network takes a sparse tensor as input and employs high-dimensional convolutions as the fundamental operator. The distinguishing characteristic of our approach is that it is able to effectively leverage local neighborhood relations together with global context even for high-dimensional data. Our network is fully-convolutional, translation-invariant, and incorporates best practices from the development of ConvNets for two-dimensional image analysis~\cite{he2016deep, batchnorm, unet}. %

To demonstrate the effectiveness and generality of our approach, we tackle various geometric pattern recognition problems. We begin with the diagnostic setting of linear subspace detection and show that our construction is effective in high-dimensional spaces and at low signal-to-noise ratios.
We then apply the presented construction to geometric pattern recognition problems that arise in computer vision, including registration of three-dimensional point sets under rigid motion (a 6D problem) and correspondence estimation between images under epipolar constraints (a 4D problem). In both settings, the problem is made difficult by the presence of outliers.

Our experiments indicate that the presented construction can reliably detect geometric patterns in high-dimensional data that is heavily contaminated by noise. It can operate in regimes where existing algorithms break down. Our approach significantly improves 3D registration performance when combined with standard approaches for 3D point cloud alignment~\cite{Choi2015, Rusu2009FPFH, Zhou2016, Zhou2018}. The presented high-dimensional convolutional network also outperforms state-of-the-art methods for correspondence estimation between images~\cite{Yi2018, Zhang2019}.
All networks and training scripts are available at~\url{https://github.com/chrischoy/HighDimConvNets}.

\section{Related Work}
\noindent\textbf{Robust model fitting.}
Fitting a geometric model to a set of observations that are contaminated by outliers is a fundamental problem that frequently arises in computer vision and related fields. The most widely used approach for robust geometric model fitting is \textit{RANdom SAmple Consensus} (RANSAC)~\cite{Fischler1981}. Due to its fundamental importance, many variants and improvements of RANSAC have been proposed over the years~\cite{Brachmann2016,Chum2005,Lebeda2012,Raguram2013,Rousseeuw1984,Tennakoon2016,Torr2002,Torr2000}.

Alternatively, algorithms for robust geometric model fitting are frequently derived using techniques from robust statistics \cite{Fitzgibbon2003,Hoseinnezhad2011,Huber2011,Zhang1998}, where outlier rejection is performed by equipping an estimator with a cost function that is insensitive to gross outliers. While the resulting algorithms are computationally efficient, they require careful initialization and optimization procedures to avoid poor local optima~\cite{Zhou2016}.

Another line of work proposes to find globally optimal solutions to the consensus maximization problem~\cite{Chin2017,Li2009,Yang2014}. However, these approaches are currently computationally too demanding for many practical applications.

\vspace{2mm}
\noindent\textbf{3D registration.}
Finding reliable correspondences between a pair of surfaces is an essential step for 3D reconstruction~\cite{Choi2015, Dai2017, Dong2019, Park2017}. The problem has been conventionally framed as an energy minimization problem which can be solved using various techniques such as branch and bound~\cite{Yang2013}, Riemannian optimization~\cite{Rosen2019}, mixed-integer programming~\cite{Izatt2017}, robust error minimization~\cite{Zhou2016}, semi-definite programming~\cite{Horowitz2014, Maron2016}, or random sampling~\cite{Choi2015}.

Recent work has begun to leverage deep networks for geometric registration~\cite{Aoki2019,Pais2019,Probst2019}. These works are predominantly based on PointNet and related architectures, which detect patterns via global pooling operations~\cite{Qi2017,Zaheer2017deep}.
In contrast, we develop a high-dimensional convolutional network that operates at multiple scales and can leverage not just global but also local geometric structure.

\vspace{2mm}
\noindent\textbf{Image correspondences.}
Yi~\etal~\cite{Yi2018} and Zhang~\etal~\cite{Zhang2019} reduce essential matrix estimation to classification of correspondences into inliers and outliers. Ranftl and Koltun~\cite{Ranftl2018} present a similar formulation for fundamental matrix estimation.
Brachmann and Rother~\cite{Brachmann2019b} propose to learn a neural network that guides hypothesis sampling in RANSAC for model fitting problems. 
Dang~\etal~\cite{Dang2018} propose a numerically stable loss function for essential matrix estimation.

All of these works employ variants of PointNets to classify inliers in an unordered set of putative correspondences. These architectures, based on pointwise MLPs, lack the capacity to model local geometric structure. In contrast, we develop convolutional networks that directly leverage neighborhood relations in the high-dimensional space of correspondences.

\section{High-Dimensional Convolutional Networks}
\label{sec:construction}

In this section, we introduce the two main building blocks of our high-dimensional convolutional network construction: generalized sparse tensors and generalized convolutions.

\subsection{Sparse Tensor and Convolution}
A tensor is a multi-dimensional array that represents high-order data. A $D$-th order tensor $\mathcal{T}$ requires $D$ indices to uniquely access its elements. We denote such indices or coordinates as $\mathbf{x} = [x^1, ..., x^D]$ and the element at the coordinate as $\mathscr{T}[\mathbf{x}]$ similar to how we access components in a matrix. Likewise, a sparse tensor is a high-dimensional extension of a sparse matrix where the majority of the elements are 0. Concretely,
\begin{equation}
   \mathscr{T}[\mathbf{x}_i] = \begin{cases}
      \mathbf{f}_i \;\; & \text{if} \; \mathbf{x}_i \in \mathcal{C} \\
      0   \;\; & \text{otherwise},
   \end{cases}
\end{equation}
where $\mathcal{C} = \{\mathbf{x}_i\;|\;\mathbf{x}_i \in \mathbb{N}^D, \mathscr{T}[\mathbf{x}_i] \neq \mathbf{0} \}^N_{i=1}$ is the set of coordinates with non-zero values, $N$ is the number of non-zero elements, and $\mathbf{f}_i$ is the non-zero value at the $i$-th coordinate.
A sparse tensor feature map is a $(D+1)$-th order tensor with $\mathbf{f}_i \in \mathbb{R}^{N_{D + 1}}$ as we use the last dimension to denote the feature dimension.
A sparse tensor has the constraint that ${\mathbf{x}_i \in \mathbb{N}^D}$. We extend the sparse tensor coordinates to integer indices ${\mathbf{x}_i \in \mathbb{Z}^D}$ and define $\mathscr{T} \in \mathbb{R}^{\aleph_0^D \times N_{D + 1}}$ where $\aleph_0$ denotes the cardinality of the integer space $|\mathbb{Z}|$ to define a generalized sparse tensor.

A convolution on this generalized sparse tensor can then be defined as a simple extension of the generalized sparse convolution~\cite{choy20194d}:
\begin{align}
\mathbf{f}^{\text{out}}_\mbf{x} & = \sum_{\mbf{i} \in \mathcal{N}^D(\mbf{x}) \cap \mathcal{C}^\text{in}} \mathbf{W}_\mbf{i} \mathbf{f}^{\text{in}}_{\mbf{x} + \mbf{i}}\quad\text{ for}\quad\mbf{x} \in \mathcal{C}^{\text{out}}, \label{eq:sparse_convolution}
\end{align}
where $\mathcal{C}^\text{in}$ and $\mathcal{C}^\text{out}$ are the set of input and output locations which is predefined by the user, $\mathbf{W_i}$ is a weight matrix, and %
$\mathcal{N}^D(\mbf{x})$ defines a set of neighbors of $\mathbf{x}$ which is defined by the shape of the convolution kernel. For example, if the convolution kernel is a hypercube of size $K$, $\mathcal{N}^D(\mbf{x}) \cap \mathcal{C}^\text{in}$ is a set of all the non-zero elements of the input sparse tensor centered at $\mathbf{x}$ within the $L_\infty$-ball of extent $K$.

\subsection{Convolutional Networks}

\begin{figure}[]
  \centering  
  \includegraphics[width=0.99\linewidth]{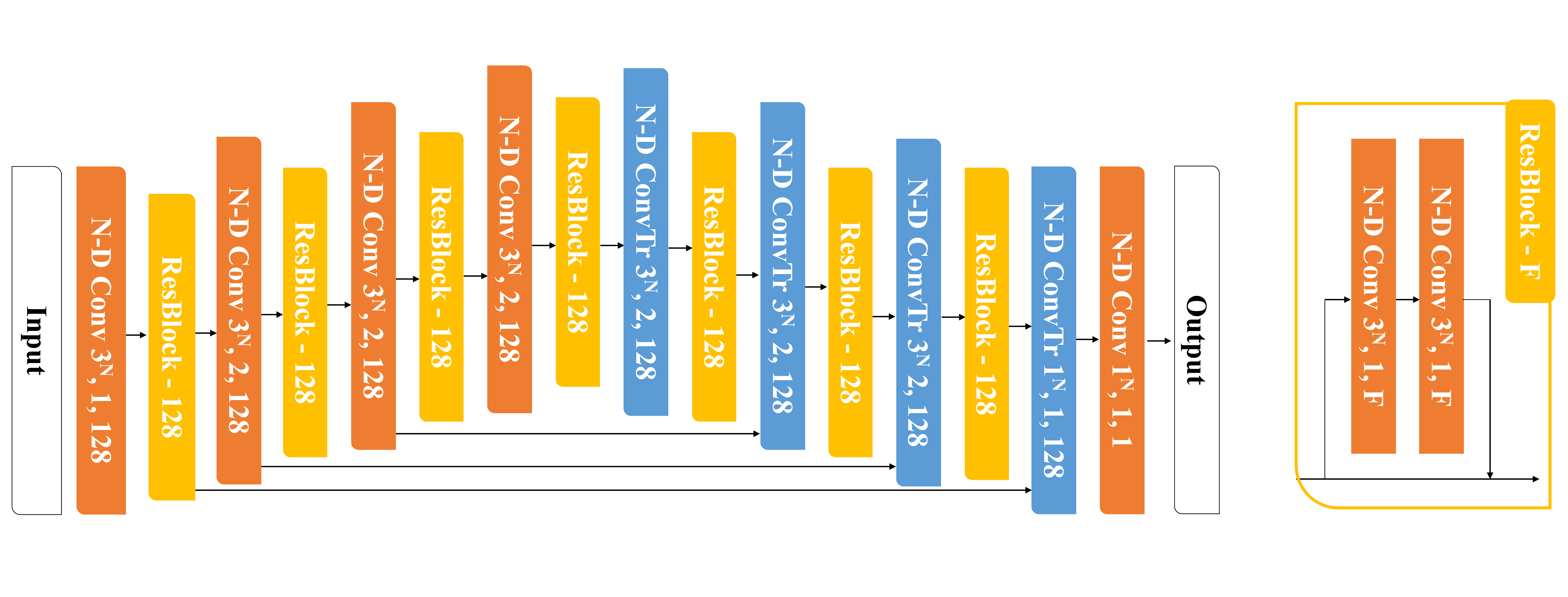}
  \caption{A generic U-shaped high-dimensional convolutional network architecture. The numbers next to each block indicate kernel size, stride, and the number of channels. The strided convolutions that reduce the resolution of activations are shifted upward to indicate different levels of resolution.
  }
  \label{fig:architecture}
\end{figure}

We design a high-dimensional fully-convolutional neural network for sparse tensors (sparse tensor networks) based on generalized convolution~\cite{choy20194d,Choy2019}.
We use U-shaped networks~\cite{unet} to capture large receptive fields while maintaining the original resolution at the final layer. The network has residual connections~\cite{he2016deep} within layers with the same resolution and across the network to speed up convergence and to recover the lost spatial resolution in the last layer. The network architecture is illustrated in Fig.~\ref{fig:architecture}.

For computational efficiency in high dimensions, we use the cross-shaped kernel~\cite{choy20194d} for all convolutions. We denote the kernel size by $K$. The cross-shaped kernel has non-zero weights only for the $K - 1$ nearest neighbors along each axis, which results in one weight parameter for the center location and $K - 1$ weight parameters for each axis. Note that a cross-shaped kernel is similar to separable convolution, where a full convolution is approximated by $D$ one-dimensional convolutions of size $K$. Both types of kernels are rank-1 approximations of the full hyper-cubic kernel $K^D$, but separable convolution requires $KD$ matrix multiplications, whereas the cross-shaped kernel requires only $(K - 1) D + 1$ matrix multiplications.

\subsection{Implementation}

We extend the implementation of Choy~\etal~\cite{choy20194d}, which supports arbitrary kernel shapes, to high-dimensional convolutional networks. To implement the sparse tensor networks, we need an efficient data structure that can generate a new sparse tensor as well as find neighbors within the sparse tensor.
Choy~\etal~\cite{choy20194d} use a hash table that is efficient for both insertion and search. We replaced the hash table with a faster and more efficient variant~\cite{robinhood}. In addition, as the neighbor search can be run in parallel, we create an iterator function that can run in parallel with OpenMP~\cite{openmp08} by dividing the table into smaller parallelization blocks.

Lastly, U-shaped networks generate hierarchical feature maps that expand the receptive field. Choy~\etal~\cite{choy20194d} use stride-$K$ convolutions with kernel size $\ge K$ to generate lower-resolution hierarchical feature maps. Still, such implementation requires iterating over at least $K^D$ elements within a hypercubic kernel as the coordinates are stored in a hash table, which results in $O(NK^D)$ complexity where $N$ is the cardinality of input. Consequently, it becomes infeasible to store the weights on the GPU for high-dimensional spaces. Instead of strided convolutions, we propose an efficient implementation of stride-$K$ sum pooling layers with kernel size $K$. Instead of iterating over all possible neighbors, we iterate over all input coordinates and round them down to multiples of $K$, which requires only $O(N)$ complexity.

\section{Geometric Pattern Recognition}
\label{sec:geometric_pattern_recognition}

Our approach uses convolutional networks to recognize geometric patterns in high-dimensional spaces. Specifically, we classify each point $\mathbf{x}_i$ in a high-dimensional dataset $\mathcal{X} = \{\mathbf{x}_i\}_{i=1}^N$ as an inlier or an outlier. We start by validating our approach on synthetic datasets of varying dimensionality and then show results on 3D registration and essential matrix estimation.

\begin{table*}[]
\centering
\caption{Line detection in high-dimensional spaces in the presence of extreme noise. All networks are trained with the cross-entropy loss for 40 epochs. Inlier ratios are listed on the left. In the 32-dimensional setting, only 7 out of 10,000 data points are inliers. The table reports Mean Squared Error (MSE), F1 score, and Average Precision (AP) of our approach (a high-dimensional ConvNet) versus baselines (PointNet variants). MSE: lower is better. F1 and AP: higher is better.}
\label{tab:line_table}
\resizebox{0.99\linewidth}{!}{
\begin{tabular}{ll|ccc|ccc|ccc|ccc}
     &          & \multicolumn{3}{|c|}{Qi~\etal~\cite{Qi2017}} & \multicolumn{3}{c|}{Zaheer~\etal~\cite{Zaheer2017deep} + BN + IN} & \multicolumn{3}{c|}{Yi~\etal~\cite{Yi2018}} & \multicolumn{3}{c}{Ours}  \\ \toprule
Dim. & Inlier Ratio      & MSE   & F1    & AP (AUC)            & MSE     & F1    & AP (AUC)                                       & MSE & F1    &  AP (AUC)                   & MSE               & F1             & AP (AUC)       \\ \midrule
4    & 15.59\%           & 1.337 & 0.025 & 0.164               & 6.33E-4 & 0.867 & 0.936                                          & 9.11E-5 & 0.981 & 0.996                   & \textbf{2.33E-5}  & \textbf{0.998} & \textbf{0.999} \\
8    &  5.54\%           & 2.369 & 0.022 & 0.065               & 0.001   & 0.891 & 0.955                                          & 2.45E-4 & 0.946 & 0.989                   & \textbf{1.64E-5}  & \textbf{0.999} & \textbf{0.999} \\
16   &  2.75\%           & 3.854 & 0.012 & 0.034               & 4.86E-4 & 0.970 & 0.992                                          & 0.002   & 0.962 & 0.986                   & \textbf{3.39E-5}  & \textbf{0.999} & \textbf{0.999} \\
24   &  0.40\%           & 5.372 & 0.021 & 0.011               & 0.676   & 0.634 & 0.691                                          & 0.775   & 0.610 & 0.674                   & \textbf{5.34E-5}  & \textbf{0.994} & \textbf{0.996} \\
32   &  0.07\%           & 6.715 & 0.012 & 6.71E-5             & -       & 0.0   & 0.295                                          & -       & 0.0   & 0.050                   & \textbf{0.010}    & \textbf{0.669} & \textbf{0.689} \\ \bottomrule
\end{tabular}
}
\end{table*}
\begin{table*}[]
\centering
\caption{Plane detection in high-dimensional spaces in the presence of extreme noise. Inlier ratios are listed on the left. In the 32-dimensional setting, fewer than 5 out of 100,000 data points are inliers. The table reports the F1 score and Average Precision (AP) of our approach (a high-dimensional ConvNet) versus baselines (PointNet variants). Higher is better.}
\label{tab:subspace_table}
\resizebox{0.8\linewidth}{!}{
\begin{tabular}{ll|cc|cc|cc|cc}
     &            & \multicolumn{2}{|c|}{Qi~\etal~\cite{Qi2017}} & \multicolumn{2}{c|}{Zaheer~\etal~\cite{Zaheer2017deep} + BN + IN} & \multicolumn{2}{c|}{Yi~\etal~\cite{Yi2018}} & \multicolumn{2}{c}{Ours}  \\ \toprule
Dim. & Inlier Ratio        & F1  & AP (AUC)                                 & F1    & AP (AUC)                                                  & F1    &  AP (AUC)                           & F1    & AP (AUC)       \\ \midrule
4    & 29.96\%    & 0.0 & 0.315                                    & 0.980 & 0.996                                                     & \textbf{0.993} & \textbf{0.999}             & 0.991 & 0.998  \\
8    &  8.07\%    & 0.0 & 0.088                                    & 0.985 & 0.999                                                     & 0.990 & 0.999                               & \textbf{0.998} & \textbf{0.999} \\
16   &  0.34\%    & 0.0 & 0.004                                    & 0.155 & 0.299                                                     & 0.182 & 0.359                               & \textbf{0.951} & \textbf{0.961} \\
24   &  0.01\%    & 0.0 & 1.61E-4                                  & 0.032 & 0.133                                                     & 0.0   & 0.081                               & \textbf{0.304} & \textbf{0.346} \\
32   &  4.64E-3\% & 0.0 & 5.56E-5                                  & 0.0   & 0.221                                                     & 0.0   & 0.023                               & \textbf{0.138} & \textbf{0.240} \\ \bottomrule
\end{tabular}
}
\end{table*}

\begin{figure*}[ht!]
  \vspace{-3mm}
  \small
  \centering  
    \begin{tabular}{cccc}
    \includegraphics[width=0.23\linewidth]{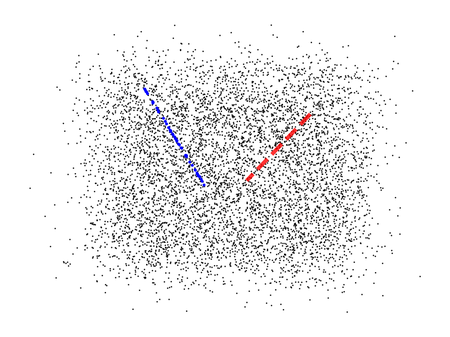} &
    \includegraphics[width=0.23\linewidth]{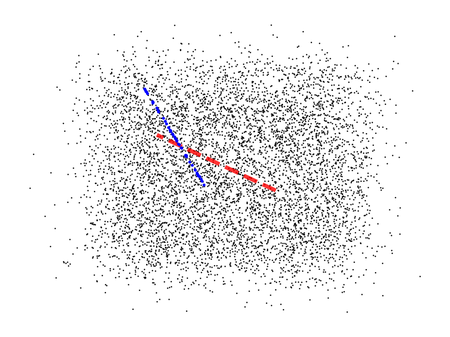} &
    \includegraphics[width=0.23\linewidth]{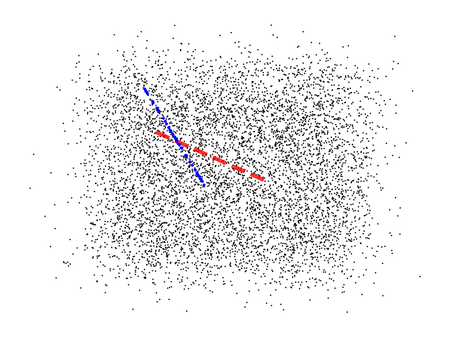} &
    \includegraphics[width=0.23\linewidth]{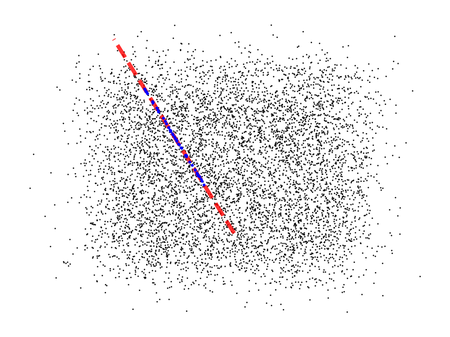} \\
    Qi~\etal~\cite{Qi2017} &
    Zaheer~\etal~\cite{Zaheer2017deep} &
    Yi~\etal~\cite{Yi2018} &
    Ours
    \end{tabular}
  \vspace{-2mm}
  \caption{16D line detection projected to a 2D plane for visualization. Black dots are noise and blue dots are samples from a line in the 16D space. The dashed red line is the prediction of the respective method. Samples from the ground-truth line (blue) are enlarged by a factor of 10 for visualization.}
  \label{fig:one_dim}
\end{figure*}

\begin{figure*}[ht]
  \small
  \centering 
    \begin{tabular}{ccc}
    \includegraphics[height=3.8cm]{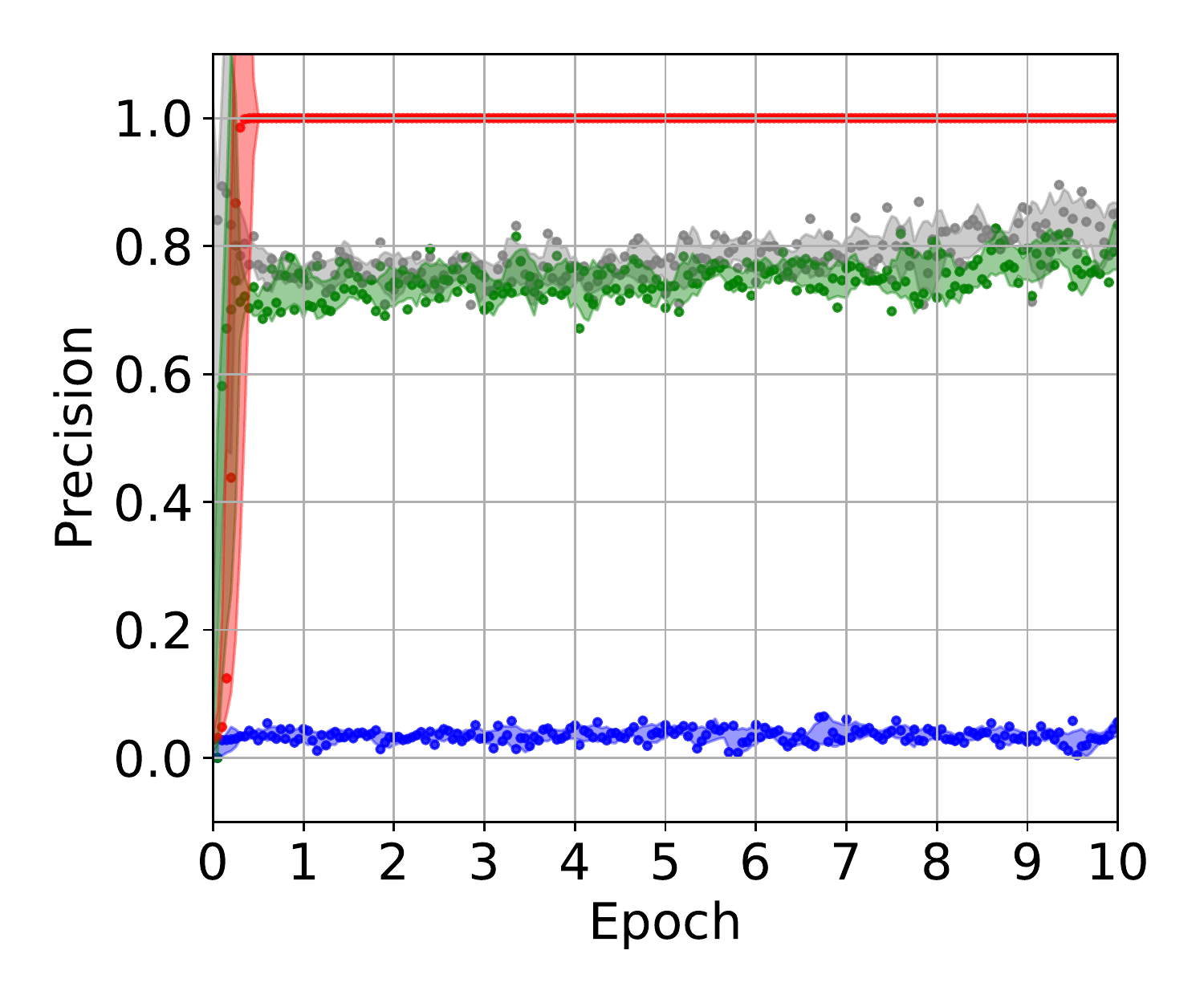} &
    \includegraphics[height=3.8cm]{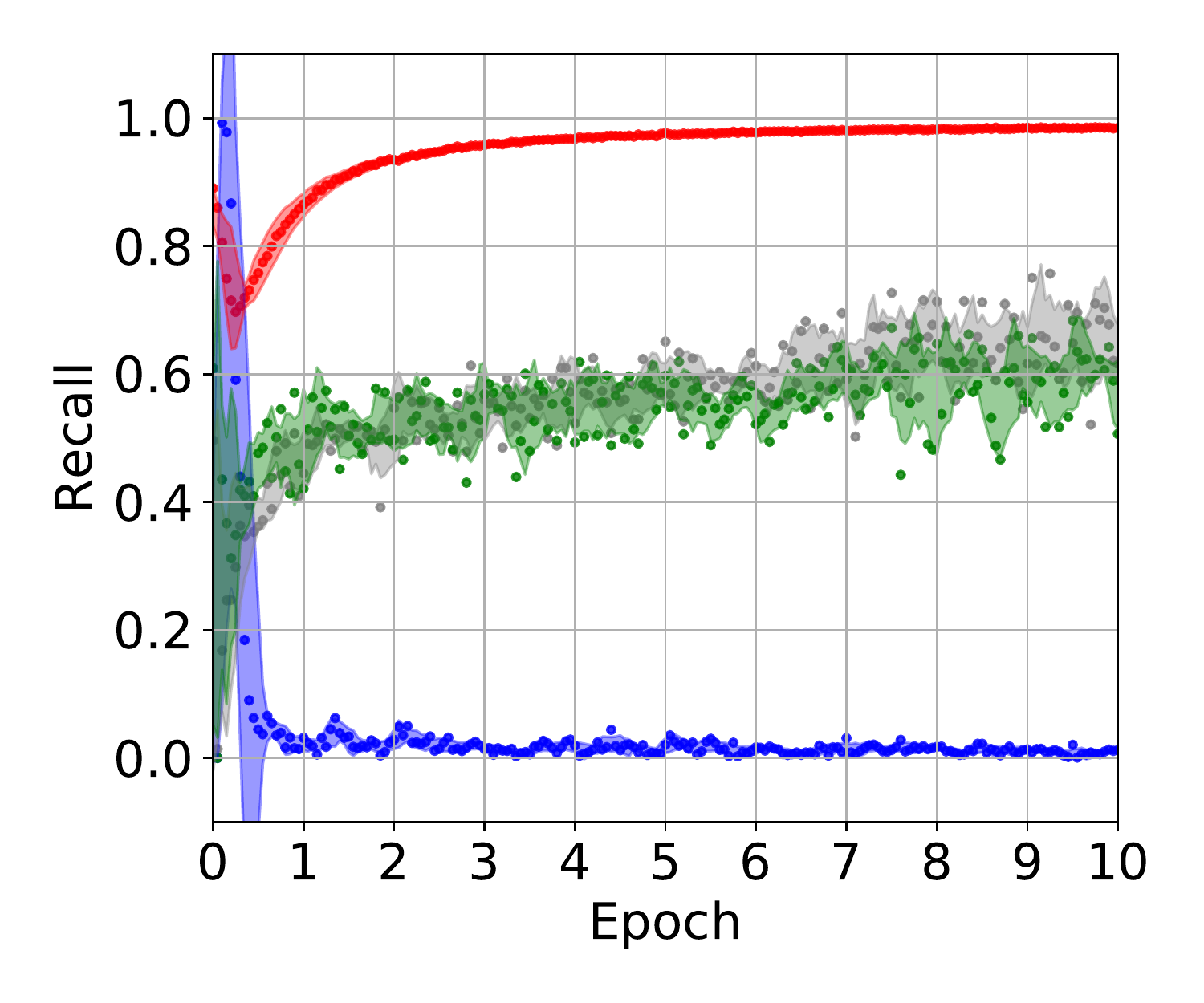} &
    \includegraphics[height=3.8cm]{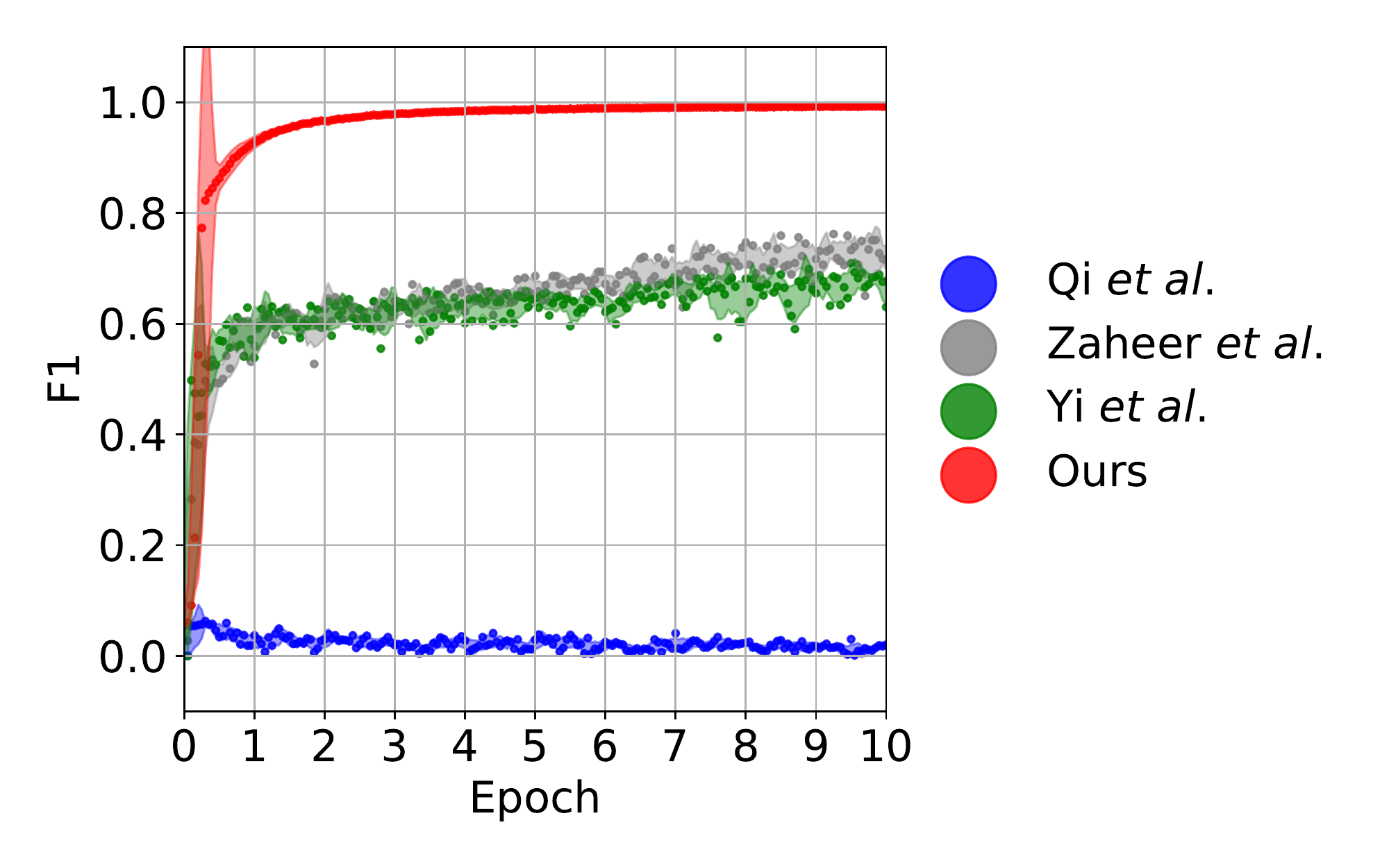} \\
    \end{tabular}
    \vspace{-1em}
  \caption{16D line detection, progression of training. We plot the running mean and standard deviation of precision, recall, and F1 score on the validation set. Our high-dimensional convolutional network quickly attains much higher accuracy than the baselines.}
  \label{fig:synth_training}
\end{figure*}

For all experiments, we first quantize the input coordinates to create a sparse tensor of order $D+1$, where the last dimension denotes the feature channels. %
The network then predicts a logit score for each non-zero element in the sparse tensor to indicate if a point is part of the geometric pattern or if it is an outlier.

\subsection{Line and Plane Detection}
\label{sec:linear_regression}

We first test the capabilities of our fully-convolutional networks on simple high-dimensional pattern recognition problems that involve detecting linear subspaces amidst noise. Our dataset consists of uniformly sampled noise from the $D$-dimensional space and a small number of samples from a line under a Gaussian noise model.
The number of outliers increases exponentially in the dimension ($O(L^D)$), while the number of inliers increases sublinearly ($O(\sqrt{LD})$), where $L$ is the extent of the domain.
Further details are given in the supplement. 
The network predicts a likelihood score for each non-zero element in the input sparse tensor, and we threshold inliers with probability $\ge 0.5$.
We estimate the line equation from the predicted inliers using unweighted least squares.

We use PointNet variants as the baselines for this experiment~\cite{Qi2017, Yi2018, Zaheer2017deep}. For Zaheer~\etal~\cite{Zaheer2017deep}, we were not able to get reasonable results with the network architecture proposed in the paper. We thus augmented the architecture with batch normalization and instance normalization layers after each linear transformation similar to Yi~\etal~\cite{Yi2018}, which boosted performance significantly.
For all experiments, we use the cross-entropy loss.
We used the same training hyperparameters, including loss, batch size, optimizer, and learning rate schedule for all approaches.

We use three metrics to analyze the performance of the networks: Mean Squared Error (MSE), F1 score, and Average Precision (AP). For the MSE, we estimate the line equation with Least Squares to fit the line to the inliers. The second metric is the F1 score, the harmonic mean of precision and recall. In many problems, F1 score is a direct indicator of the performance of a classifier, and we also found a strong correlation between F1 score and the mean squared error. The final metric we use is average precision (AP), which measures the area under the precision-recall curve. We report the results in Tab.~\ref{tab:line_table} and provide qualitative examples in Fig.~\ref{fig:one_dim}. Tab.~\ref{tab:line_table} lists the inlier ratio to indicate the difficulty of each task.

In a second experiment, we create another synthetic dataset where the inlier pattern is sampled from a plane spanned by two vectors: $\{c_1 \mathbf{v}_1 + c_2 \mathbf{v}_2 + \mathbf{c}\;|\;c_1, c_2 \in \mathbb{R}\}$. The two basis vectors are sampled uniformly from the $D$-dimensional unit hypercube. We use the same training procedure for our network and the baselines, and report the results in Tab.~\ref{tab:subspace_table}.

We found that the convolutional network is more robust to noise in high-dimensional spaces than the PointNet variants. In addition, the convolutional network training converges quickly, as shown in Fig.~\ref{fig:synth_training}, which is a further indication that the architecture can effectively leverage the structure of the data.

\subsection{3D Registration}
\label{sec:hypersurface}

A typical 3D registration pipeline consists of 1) feature extraction, 2) feature matching, 3) match filtering, and 4) global registration. In this section, we show that in the \textit{match filtering} stage the correct (inlier) correspondences form a 6-dimensional geometric structure. We then extend our geometric pattern recognition networks to identify inlier correspondences in this 6-dimensional space.

\begin{figure}[t]
  \centering
  \small
    \includegraphics[width=0.50\linewidth]{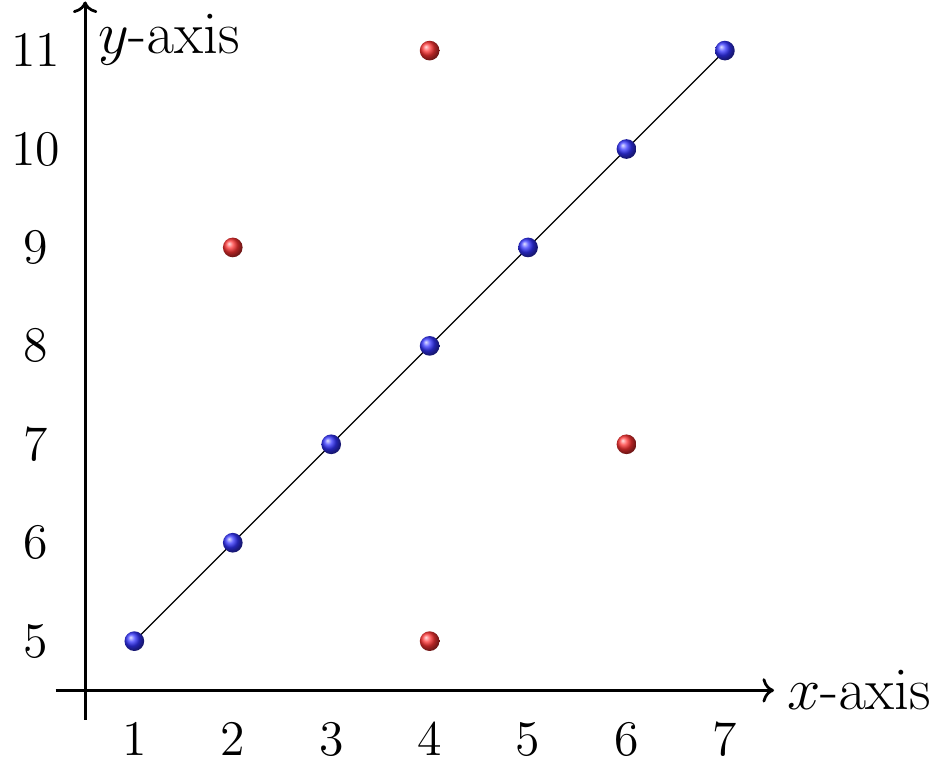}
  \caption{The set $\mathcal{Y}$ is rigid translation of the set $\mathcal{X} = [1, ..., 7]$, $\mathcal{Y} = \{x + 4| x \in \mathcal{X}\}$. 
  The ordered pairs of correspondences (blue) form a line segment while outliers (red) form random noise outside the line.}
  \label{fig:correspondence_diagram}
\end{figure}

Let $\mathcal{X}$ be a set of points sampled from a 3D surface, $\mathcal{X} = \{\mathbf{x}_i\;|\;\mathbf{x}_i \in \mathbb{R}^3\}_{i=1}^N$, and let $\mathcal{X}'$ be a subset of $\mathcal{X}$ that went through a rigid transformation $T$, $\mathcal{X}' = \{T(\mathbf{x})\;|\;\mathbf{x} \in \mathcal{S}, \mathcal{S} \subseteq \mathcal{X} \}$. For example, $\mathcal{X}'$ could be a 3D scan from a different perspective that has an overlap with $\mathcal{X}$.
We denote a correspondence between points $\mathbf{x}_i \in \mathcal{X}$ and $\mathbf{x}'_j \in \mathcal{X}'$ as $\mathbf{x}_i \leftrightarrow \mathbf{x}'_j$.
When we form an ordered pair $(\mathbf{x}_i, \mathbf{x}'_j) \in \mathbb{R}^6$, the ground truth correspondences satisfy $T(\mathbf{x}) = \mathbf{x}'$ along the common 3D geometry $\mathcal{S}$ whereas an incorrect correspondence implies $T(\mathbf{x}) \neq \mathbf{x}'$.
For example, in Fig.~\ref{fig:correspondence_diagram}, we visualize the ordered pairs from 1-dimensional sets. Note that the inliers follow the geometry of the inputs and form a line segment. 
Similarly, the geometry $(\mathbf{x}, \mathbf{x}') \in \mathbb{R}^6$ or $(\mathbf{x}, T(\mathbf{x}))$ for $\mathbf{x} \in \mathcal{S}$ forms a surface in 6-dimensional space.
\begin{theorem}
The 6D representation of ground-truth 3D correspondences $(\mathbf{x}, \mathbf{x}')$ lies on the intersection of three hyperplanes where each hyperplane is a row of the following block equation $\begin{bmatrix} R & - I \end{bmatrix} \begin{bmatrix}\mathbf{x} \\ \mathbf{x}' \end{bmatrix} + \mathbf{t} = \mathbf{0}$.
\end{theorem}
\textbf{Proof:} A ground-truth 3D correspondence, $\mathbf{x} \leftrightarrow \mathbf{x}'$, must satisfy $R\mathbf{x} + \mathbf{t} = \mathbf{x}'$. We move $\mathbf{x}'$ to the LHS and convert both $\mathbf{x}$ and $\mathbf{x}'$ to the 6D representation to get three hyperplane equations.
\hfill\ensuremath{\square}
\begin{corollary}
The rank of the block matrix, $\begin{bmatrix} R & - I \end{bmatrix}$, is 3 since $R$ and $I$ are orthogonal matrices. Thus, all hyperplanes defined by the block matrix intersect each other and the 6D inlier correspondences form a 3D plane since the solution of the intersection of three hyperplanes in the 6D space, $\begin{bmatrix} R & - I \end{bmatrix} \begin{bmatrix}\mathbf{x} \\ \mathbf{x}' \end{bmatrix} + \mathbf{t} = \mathbf{0}$, is a 3D plane.
\end{corollary}
 
We can thus use our high-dimensional convolutional network construction to segment the 6-dimensional set of correspondences into inliers and outliers by estimating the inlier likelihood for each correspondence.

\vspace{2mm}
\noindent\textbf{Network.}
We use a 6-dimensional instantiation of the U-shaped convolutional network presented in Sec.~\ref{sec:construction}. As the dimensionality is manageable, we use hypercubic kernels.
The network takes an order-6 sparse tensor whose coordinates are correspondences $(\mathbf{x}_i, \mathbf{x}'_j)\in \mathbb{R}^6$. We discretize the coordinates with the voxel size used to extract features. Our baseline is Yi~\etal~\cite{Yi2018}, which takes dimensionless mean-centered correspondences without discretization. We train the networks to predict the inlier probability of each correspondence with the balanced cross-entropy loss.

\vspace{2mm}
\noindent\textbf{Dataset.} We use the 3DMatch dataset for this experiment~\cite{Zeng20163dmatch}. The 3DMatch dataset is a composition of various 3D scan datasets~\cite{Glocker2013, xiao2013sun3d, Zeng20163dmatch} and thus covers a wide range of scenes and different types of 3D cameras. We integrate RGB-D images to form fragments of the scenes following~\cite{Zeng20163dmatch}. During training, we randomly rotate each scene on the fly to augment the dataset.
We use a popular hand-designed feature descriptor, FPFH~\cite{Rusu2009FPFH}, to compute correspondences. Note, however, that our pipeline is agnostic to the choice of feature and can also be used with learned features~\cite{Choy2019}.

\begin{table*}[]
\centering
\small
  \caption{Pairwise registration on 3DMatch test scenes with 2.5cm downsampling. Translation Error (TE), Rotation Error (RE), success rate. Registration is considered successful if TE $< 30$cm and RE $< 15{}^\circ$.}
  \label{tab:three_dim_2_5cm}
  \resizebox{0.95\linewidth}{!}{
  \begin{tabular}{@{}lc|ccc|ccc||ccc|ccc@{}}
    \toprule
                &         & \multicolumn{3}{c|}{FPFH + FGR}    & \multicolumn{3}{c||}{FPFH + Ours + FGR}    & \multicolumn{3}{c|}{FPFH + RANSAC}  & \multicolumn{3}{c}{FPFH + Ours + RANSAC}  \\
                & Inlier Ratio     & TE & RE & Succ. Rate               & TE   & RE   & Succ. Rate                  & TE & RE & Succ. Rate & TE & RE & Succ. Rate \\ \midrule
    Kitchen     & 1.62\%  & 10.98 & 4.99 & 37.15               & 5.68 & 2.21 & 65.61                       & 6.25 & 2.17 & 44.47                 & \textbf{5.90} & \textbf{1.98} & \textbf{69.57}   \\
    Home 1      & 2.71\%  & 11.12 & 4.40 & 45.51               & 6.52 & 2.08 & \textbf{80.77}                       & 7.07 & 2.19 & 61.54                 & \textbf{6.00} & \textbf{1.87} & 80.13   \\
    Home 2      & 2.83\%  & 9.61 & 3.83 & 36.54                & 7.13 & 2.56 & 64.42                       & \textbf{6.47} & \textbf{2.40} & 50.00                 & 7.86 & 2.56 & \textbf{69.71}   \\
    Hotel 1     & 1.35\%  & 12.31 & 5.09 & 33.19               & 7.95 & 2.65 & 76.11                       & 7.48 & 2.75 & 48.67                 & \textbf{7.38} & \textbf{2.38} & \textbf{80.09}   \\
    Hotel 2     & 1.54\%  & 12.27 & 5.22 & 25.00               & 7.86 & 2.56 & 69.23                       & 9.54 & 3.18 & 47.12                 & \textbf{6.40} & \textbf{2.25} & \textbf{70.19}   \\
    Hotel 3     & 1.59\%  & 13.52 & 7.04 & 27.78               & \textbf{5.39} & \textbf{1.99} & 72.22                       & 5.91 & 2.46 & 59.26                 & 5.85 & 2.36 & \textbf{81.48}   \\
    Study       & 0.87\%  & 16.10 & 6.01 & 16.78               & 9.61 & 2.64 & 53.42                       & 10.05 & 3.01 & 30.48                & \textbf{8.51} & \textbf{2.23} & \textbf{56.16}   \\
    Lab         & 1.59\%  & 10.48 & 4.80 & 42.86               & 7.69 & 2.44 & 61.04                       & 8.01 & 2.31 & 45.45                 & \textbf{6.64} & \textbf{2.12} & \textbf{68.83}   \\ \midrule
    Average     &         & 12.05 & 5.17 & 33.10               & 7.23 & 2.39 & 67.85                       & 7.60 & 2.56 & 48.37                 & \textbf{6.82} & \textbf{2.22} & \textbf{72.02}   \\ \midrule
  \end{tabular}
  }
\end{table*}

\begin{figure*}[t]
  \centering
  \small
      \begin{tabular}{c}
    \includegraphics[width=0.90\linewidth]{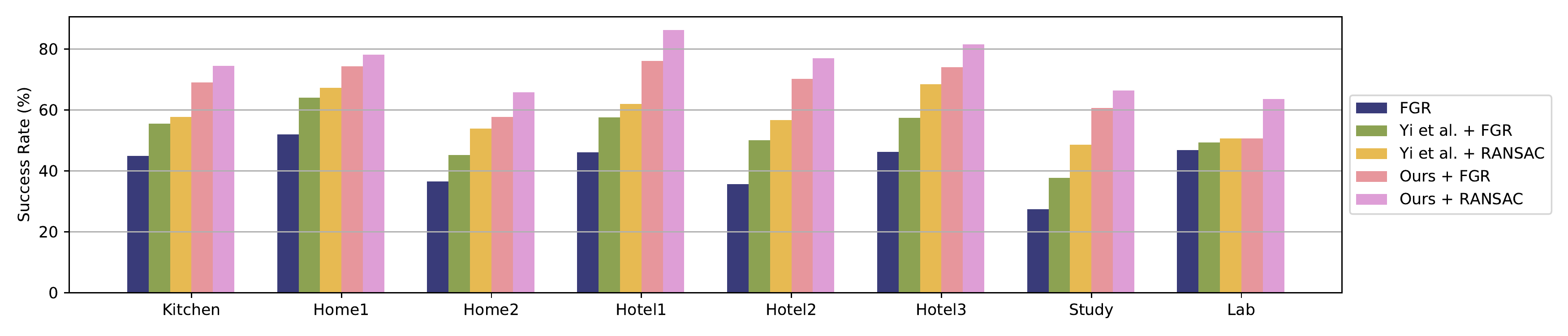} 
    \end{tabular}
  \caption{The success rate of baseline methods and ours on the 3DMatch benchmark~\cite{Zeng20163dmatch} with 5cm voxel size. \textit{FGR} denotes registration with FPFH~\cite{Rusu2009FPFH} and Zhou~\etal~\cite{Zhou2016}, \textit{Yi~\etal + X} denotes FPFH filtering with Yi~\etal~\cite{Yi2018} and registration with X, and \textit{Ours + X} denotes our method for filtering followed by registration with X.
  } 
  \label{fig:3dmatch_succ_rate}
\end{figure*}
\begin{figure*}[ht!]
\small
  \centering
    \resizebox{0.92\linewidth}{!}{
      \begin{tabular}{cc|cc}
    \includegraphics[width=0.21\linewidth]{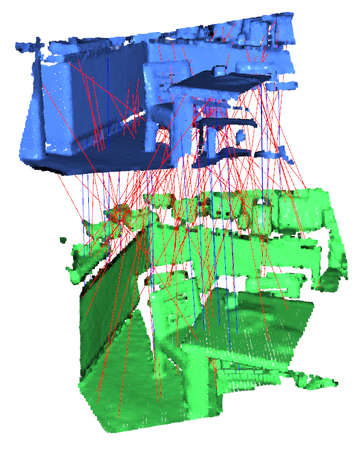} &
    \includegraphics[width=0.21\linewidth]{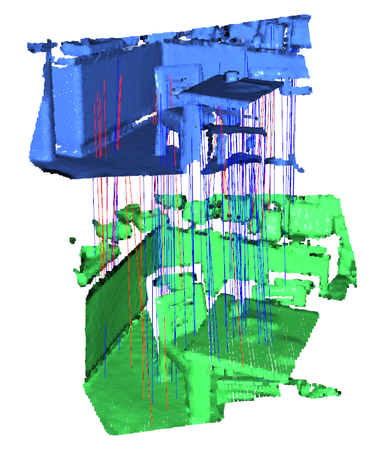} &
    \includegraphics[width=0.15\linewidth]{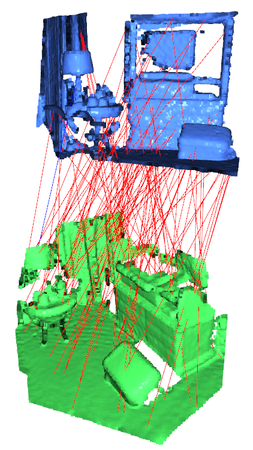} &
    \includegraphics[width=0.15\linewidth]{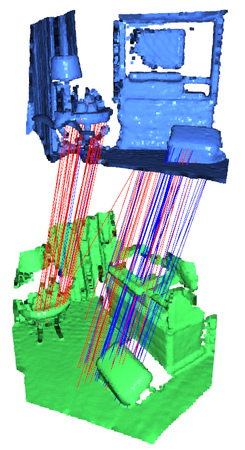} \\
    \multicolumn{2}{c|}{Kitchen} & \multicolumn{2}{c}{Bedroom} \\
    \includegraphics[width=0.21\linewidth]{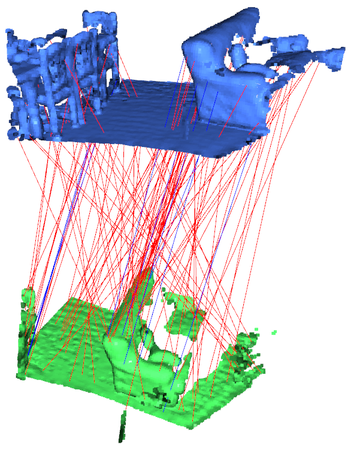} &
    \includegraphics[width=0.21\linewidth]{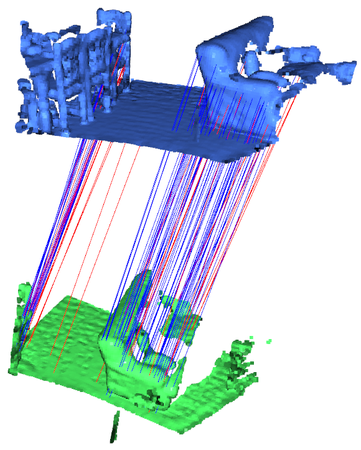} &
    \includegraphics[width=0.21\linewidth]{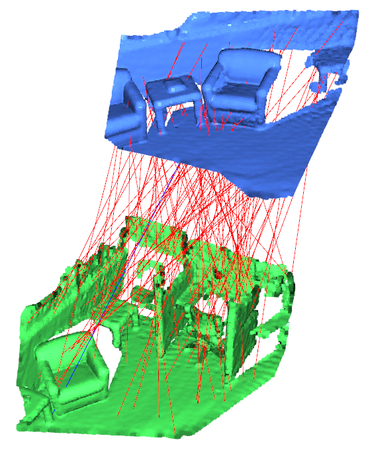} &
    \includegraphics[width=0.21\linewidth]{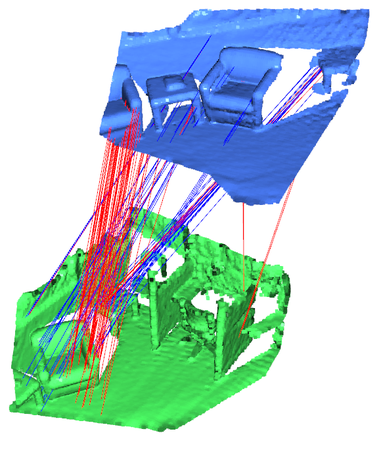} \\
    \multicolumn{2}{c}{Livingroom} & \multicolumn{2}{c}{Study} \\
    \end{tabular}
    }
  \caption{Visualization of color-coded correspondences before and after outlier filtering (Sec.~\ref{sec:hypersurface}). For each pair, we visualize 100 random correspondences from the candidate set on the left and 100 random correspondences after outlier pruning on the right. Red lines are outlier correspondences and blue lines are inlier correpondences. On the bottom right pair, there are two identical chairs. The average inlier ratio is 1.76\%.}
  \label{fig:3dmatch_vis}
\end{figure*}

We follow the standard procedures in the 3D registration literature to generate candidate correspondences. First, since 3D scans often exhibit irregular densities, we resample the input point clouds using a voxel grid to produce a regular point cloud. We use voxel sizes of 2.5cm and 5cm for our experiments. %
Next, we compute FPFH features and find the nearest neighbor for each point in feature space to form correspondences. 
The correspondences obtained from this procedure often exhibit a very low inlier ratio, as little as 0.87\% with a 2.5cm voxel size.
Among these correspondences, we regard $\mathbf{x}\leftrightarrow \mathbf{x}'$ as an inlier if it satisfies $\|\mathbf{T}(\mathbf{x})-\mathbf{x}'\|_2 <\tau$ and all others as outliers. We set $\tau$ to be two times the voxel size.

Finally, we use a registration method to convert the filtered correspondences into a final registration result. We show results with two different registration methods. The first is Fast Global Registration~\cite{Zhou2016}, which directly minimizes a robust error metric. The second is a variant of RANSAC~\cite{Fischler1981} that is specialized to 3D registration~\cite{Zhou2018}.

\vspace{2mm}
\noindent\textbf{Evaluation.}
We use three standard metrics to evaluate registration performance: rotation error, translation error, and success rate. The rotation error measures the absolute angular deviation from the ground truth rotation $\hat{\textbf{R}}$, $\arccos \frac{\text{Tr}(\hat{\textbf{R}}^T\textbf{R}) - 1}{2}$. Similarly, the translation error measures the deviation of the translation $\| \hat{\mathbf{t}} - \mathbf{t}\|_2$. When we report these metrics, we exclude alignments that exceed a threshold following~\cite{Choy2019} since the results of the registration methods~\cite{Zhou2016, Fischler1981} can be arbitrarily bad when registration fails.
Finally, the success rate is the ratio of registrations that were successful; registration is considered successful if both rotation and translation errors are within the respective thresholds. For all experiments, we use a rotation error of 15 degrees and a translation error of 30cm as the thresholds.

Tab.~\ref{tab:three_dim_2_5cm} shows the 3D registration pipelines with and without our network to filter the outliers. Note that for FGR~\cite{Zhou2016}, we observe a considerable improvement with our network since FGR assumes more accurate correspondences as inputs. The improvement is smaller with RANSAC, since it is more robust to high outlier rates.
Although the inlier ratio is as low as 1\% for many 3D scene pairs, our network generates very accurate predictions. Similar to the linear regression experiments in Fig.~\ref{fig:synth_training}, we find that the network converges very quickly.
We compare to the model of Yi~\etal~\cite{Yi2018} in Fig.~\ref{fig:3dmatch_succ_rate} with 5cm voxel size to study the robustness of the 6-dimensional convolutional network to voxel size and find that the convolutional network improves the registration success rate significantly even for the higher inlier ratio seen with a 5cm discretization resolution.
Qualitatively, our network accurately filters out outliers even in the presence of extreme noise (Fig.~\ref{fig:3dmatch_vis}).

\subsection{Filtering Image Correspondences}

In this section, we apply the high-dimensional convolutional network to image correspondence inlier detection. In the projective space $\mathbb{P}^2$, an inlier correspondence ${\mathbf{u} \leftrightarrow \mathbf{u}'}$ must satisfy $\mathbf{u}'^\top \mathbf{E} \mathbf{u} = 0$, where $\mathbf{E}$ is an essential matrix, $\mathbf{u}$ denotes a normalized homogeneous coordinate $\mathbf{u} = \mathbf{K}^{-1} \mathbf{x}$, $\mathbf{x}$ is the corresponding homogeneous image coordinate, and $\mathbf{K}$ is the camera intrinsic matrix.
When we expand $\mathbf{u}'^\top \mathbf{E} \mathbf{u} = 0$,
we get $u'^{1}Au^1 + u'^{2}Bu^1 + u'^{1}Cu^2 + u'^{2}Du^2 + Eu'^1 + Fu'^2 + Gu^1 + Hu^2 + I = 0$,
which is a quadrivariate quadratic function. If there is a real-valued solution, there are infinitely many solutions that form either an ellipse (sphere), a parabola, or a hyperbola. These are known as conic sections. Thus a set of ground-truth image correspondences will form a hyper conic section in 4-dimensional space. We use a convolutional network to predict the likelihood that a correspondence is an inlier.

\vspace{2mm}
\noindent\textbf{Dataset: YFCC100M.}
We use the large-scale photo-tourism dataset YFCC100M~\cite{thomee2016yfcc100m} for the experiment. The dataset contains 100M Flicker images of tourist hot-spots with metadata, which is curated into 72 locations with camera extrinsics estimated using SfM~\cite{heinly2015reconstructing}.
We follow Zhang~\etal~\cite{Zhang2019} to generate a dataset and use 68 locations for training and the others for testing. We filtered any image pairs that have fewer than 100 overlapping 3D points from SfM to guarantee non-zero overlap between images.

We use SIFT features~\cite{Lowe2004SIFT} to create correspondences and label a correspondence to be a ground-truth inlier if the symmetric epipolar distance of the correspondence is below a certain threshold using the provided camera parameters, \ie, 
\begin{equation}
\label{eq:symmetric}
    \Big(\frac{r^2}{l_1^2+l_2^2} + \frac{r^2}{l_1'^2+l_2'^2}\Big) < \tau,
\end{equation}
where $l = \mathbf{u}'^\top \mathbf{E} = (l_1, l_2, l_3)$ is a homogeneous line, $l'=\mathbf{E}\mathbf{u}$, and
$r = \mathbf{u}'^\top \mathbf{E} \mathbf{u}$.

\vspace{2mm}
\noindent\textbf{Network.}
We convert a set of candidate image correspondences into an order-5 sparse tensor with four spatial dimensions and vectorized features. The coordinates are defined as normalized image coordinates $\mathbf{u}$.
We define integer coordinates by discretizing the normalized image coordinates with quantization resolution 0.01 and additionally use the normalized coordinates as features.
We use state-of-the-art baselines on the YFCC dataset and two variants of convolutional networks for this task. The first variant is a U-shaped convolutional network (Ours); the second network is a ResNet-like network with spatial correlation modules (Ours + SC). Spatial correlation modules~\cite{Zhang2019} are blocks of shared MLPs that encode global context from a set of correspondences. 

Note that unlike the FPFH descriptor~\cite{Rusu2009FPFH}, which extracts features densely, SIFT features are extracted from only a few keypoints sparsely in the image. In the 4-dimensional space of correspondences, the sparsity gets even worse as the volume increases multiplicatively while the number of points (correspondences) stay the same.
Such sparsity leads to fewer neighbors in the high-dimensional space, which results in the degeneration of the convolution to a multi-layer perceptron. Our convolutional networks remain effective in this setting, but their distinctive ability to leverage local geometric structure is not strongly utilized.
To increase the density of the correspondences in the 4-dimensional space, we use a dense fully convolutional feature UCN~\cite{UCN2016}. We train the UCN on the YFCC100M training set for 100 epochs and we follow the preprocessing stage outlined in the open-source version~\cite{openucn} to create correspondences. Training and testing on UCN follow the same standard procedure.
We use the balanced cross-entropy loss~\cite{Yi2018} for all experiments.

\begin{table*}[ht!]
\centering
\caption{Classification scores on the YFCC100M test set. We consider a correspondence a true positive if the symmetric epipolar distance is below $10^{-4}$ and the prediction confidence is above 0.5.}
\label{tab:two_dim_baseline}
\resizebox{0.85\linewidth}{!}{
\begin{tabular}{@{}l||ccc|ccc|ccc|ccc|ccc@{}}
\toprule
        & \multicolumn{3}{c|}{Yi~\etal~\cite{Yi2018}} & \multicolumn{3}{c|}{Zhang~\etal~\cite{Zhang2019}} & \multicolumn{3}{c|}{Ours}       & \multicolumn{3}{c|}{Ours + SC} & \multicolumn{3}{c}{Ours + UCN} \\
                     & Prec. & Recall & F1            & Prec. & Recall & F1                      & Prec. & Recall & F1             & Prec. & Recall & F1    & Prec. & Recall & F1 \\ \midrule
\textsc{Buckingham}  & 0.497 &  0.772 &  0.605        & 0.486 & 0.889 & 0.629                    & 0.535 & 0.822 & 0.648           & 0.611 & 0.835 & 0.705  & \textbf{0.769} & \textbf{0.892} & \textbf{0.826}\\
\textsc{Notre dame}  & 0.581 &  0.894 &  0.705        & 0.629 & 0.951 & 0.757                    & 0.647 & 0.915 & 0.758           & 0.721 & 0.929 & 0.812  & \textbf{0.844} & \textbf{0.954} & \textbf{0.896}\\
\textsc{Reichtag}    & 0.747 &  0.877 &  0.807        & 0.734 & 0.917 & 0.815                    & 0.695 & 0.911 & 0.789           & 0.769 & 0.897 & 0.827  & \textbf{0.856} & \textbf{0.937} & \textbf{0.895}\\
\textsc{Sacre coeur} & 0.658 &  0.871 &  0.750        & 0.662 & 0.948 & 0.780                    & 0.632 & 0.917 & 0.748           & 0.718 & 0.932 & 0.811  & \textbf{0.855} & \textbf{0.948} & \textbf{0.899} \\ \midrule
Average              & 0.621 &  0.854 &  0.717        & 0.628 & 0.926 & 0.745                    & 0.628 & 0.891 & 0.736           & 0.704 & 0.898 & 0.789  & \textbf{0.830} & \textbf{0.933} & \textbf{0.879} \\ \bottomrule
\end{tabular}
}
\end{table*}
\newcommand\yfcch{0.19}
\newcommand{\STAB}[1]{\begin{tabular}{@{}c@{}}#1\end{tabular}}
\begin{figure*}[ht!]
  \small
  \centering  
  \resizebox{0.995\linewidth}{!}{
  \begin{tabular}{cccccc}
    \multicolumn{2}{c}{\textsc{Buckingham}} & 
    \multicolumn{2}{c}{\textsc{Notre Dame}} &
    \multicolumn{2}{c}{\textsc{Sacre Coeur}} 
    \\
    \includegraphics[height=\yfcch\textheight]{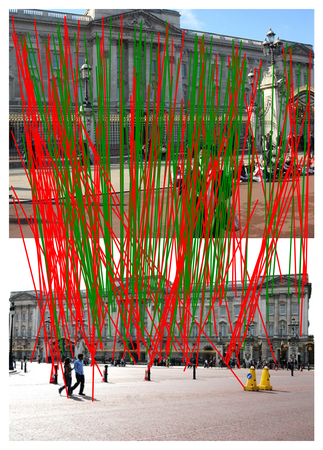} &
    \includegraphics[height=\yfcch\textheight]{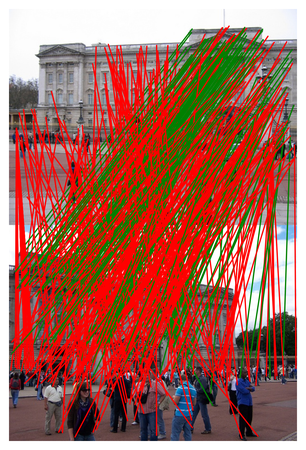} &
    \includegraphics[height=\yfcch\textheight]{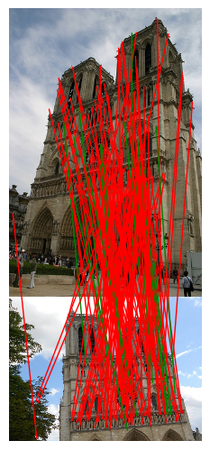} &
    \includegraphics[height=\yfcch\textheight]{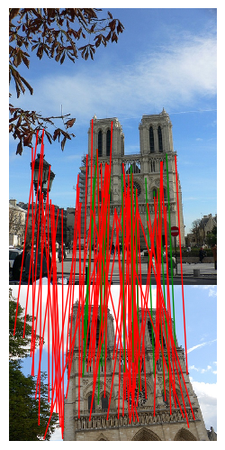} &
    \includegraphics[height=\yfcch\textheight]{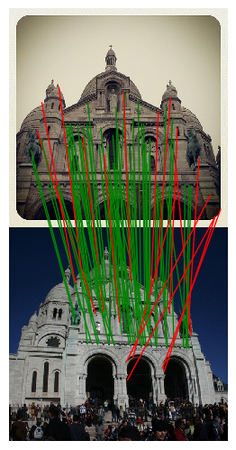} &
    \includegraphics[height=\yfcch\textheight]{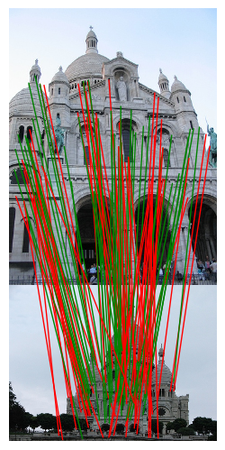} \\

    \includegraphics[height=\yfcch\textheight]{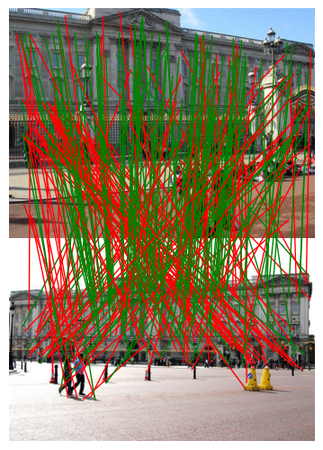} &
    \includegraphics[height=\yfcch\textheight]{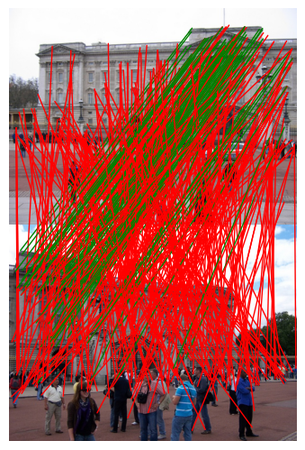} &
    \includegraphics[height=\yfcch\textheight]{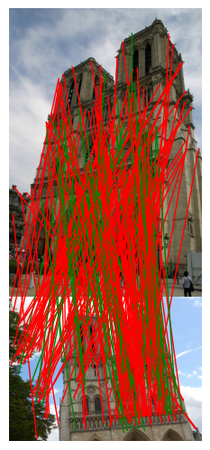} &
    \includegraphics[height=\yfcch\textheight]{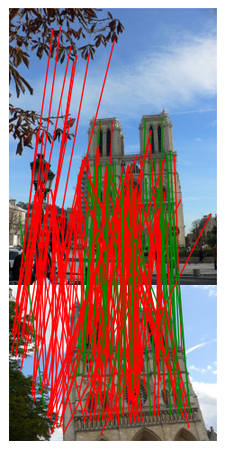} &
    \includegraphics[height=\yfcch\textheight]{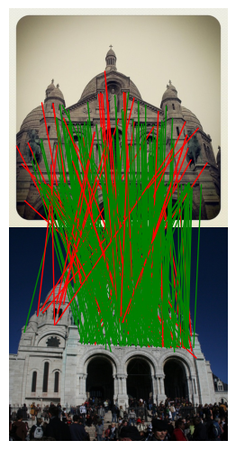} &
    \includegraphics[height=\yfcch\textheight]{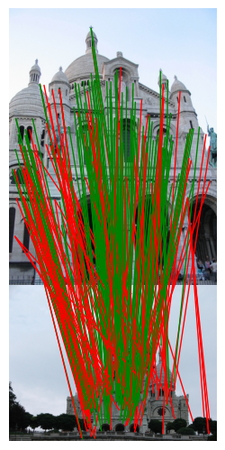} \\
    
    \includegraphics[height=\yfcch\textheight]{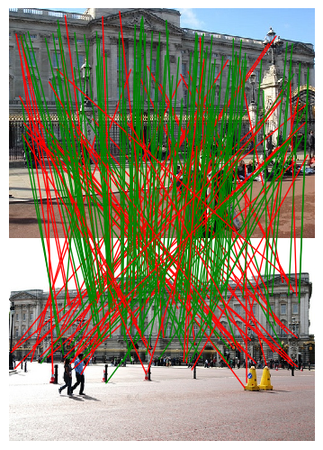} &
    \includegraphics[height=\yfcch\textheight]{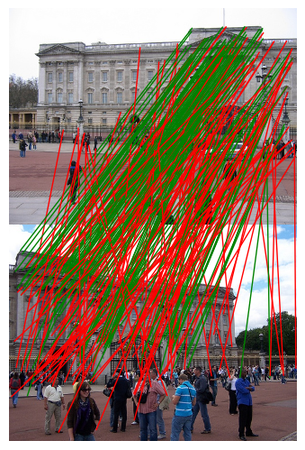} &
    \includegraphics[height=\yfcch\textheight]{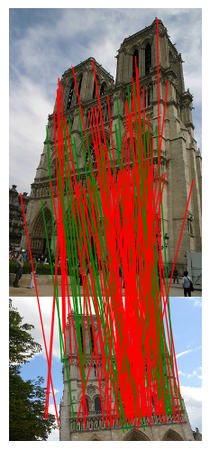} &
    \includegraphics[height=\yfcch\textheight]{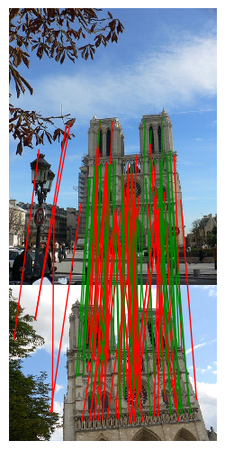} &
    \includegraphics[height=\yfcch\textheight]{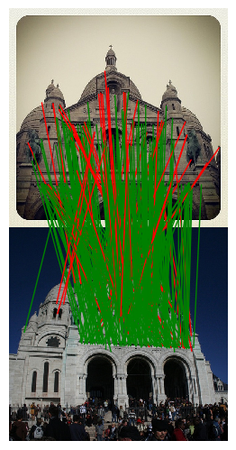} &
    \includegraphics[height=\yfcch\textheight]{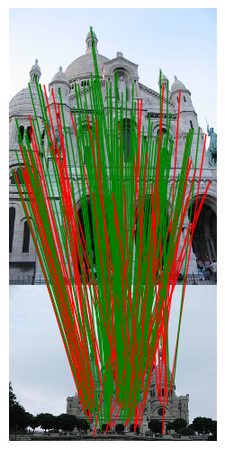} \\
    
    \includegraphics[height=\yfcch\textheight]{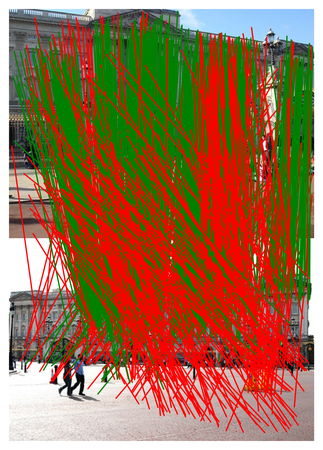} &
    \includegraphics[height=\yfcch\textheight]{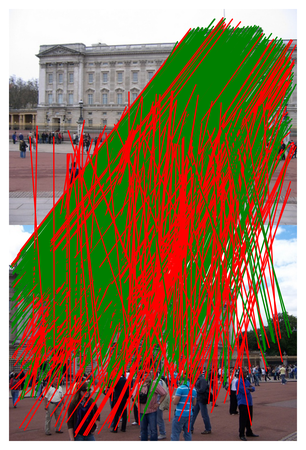} &
    \includegraphics[height=\yfcch\textheight]{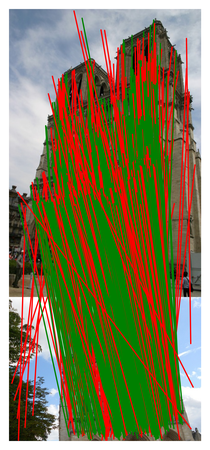} &
    \includegraphics[height=\yfcch\textheight]{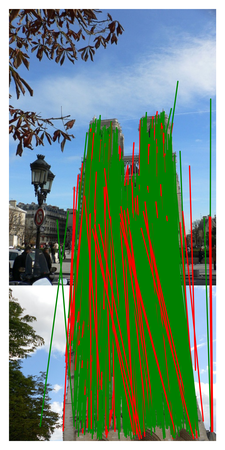} &
    \includegraphics[height=\yfcch\textheight]{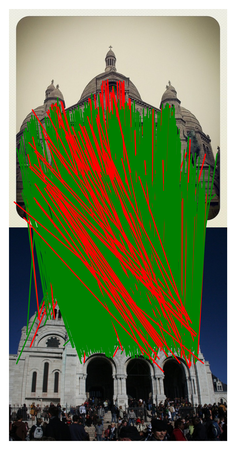} &
    \includegraphics[height=\yfcch\textheight]{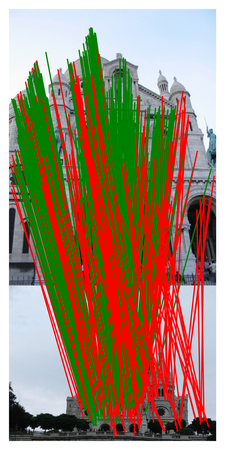} \\
  \end{tabular}
  }
  \caption{Matching results using Yi~\etal~\cite{Yi2018} (first row), Zhang~\etal~\cite{Zhang2019} (second row), Ours + SC (third row) and Ours + UCN (last row). We visualize correspondences with inlier probability above 0.5. We color a correspondence green if it is a true positive (symmetric epipolar distance smaller than $10^{-4}$) and red otherwise.}
  \label{fig:yfcc_comparison}
\end{figure*}

\vspace{2mm}
\noindent\textbf{Evaluation.}
We use precision, recall, and F1 score to evaluate correspondence classification accuracy. We use $\tau = 10^{-4}$ for the distance threshold to define ground truth-correspondences. Our network predicts a correspondence to be accurate if its inlier probability prediction is above 0.5. %
We provide quantitative results in Tab.~\ref{tab:two_dim_baseline} and qualitative results in Fig.~\ref{fig:yfcc_comparison}.
Our approachs outperform the PointNet variants~\cite{Yi2018, Zhang2019} as measured by the F1 score.

\section{Conclusion}
Many interesting problems in computer vision involve geometric patterns
in high-dimensional spaces. We proposed high-dimensional
convolutional networks for geometric pattern recognition. We presented a fully-convolutional
network architecture that is efficient and able to find patterns in high-dimensional 
data even in the presence of severe noise.
We validated the efficacy of our approach on tasks such as line and plane detection, 3D registration,
and geometric filtering of image correspondences.

\section*{Acknowledgements}

This work was partially supported by a National Research Foundation of Korea (NRF) grant and funded by the Korean government (MSIT) (No. 2020R1C1C1015260).

\balance
{\small
\bibliographystyle{ieee_fullname}
\bibliography{egbib}

\begin{thebibliography}{10}\itemsep=-1pt

\bibitem{robinhood}
Martin Ankerl.
\newblock Robin hood hashing.
\newblock \url{https://github.com/martinus/robin-hood-hashing}, 2019.

\bibitem{Aoki2019}
Yasuhiro Aoki, Hunter Goforth, Rangaprasad Arun~Srivatsan, and Simon Lucey.
\newblock {PointNetLK}: Robust \& efficient point cloud registration using
  pointnet.
\newblock In {\em CVPR}, 2019.

\bibitem{Brachmann2016}
Eric Brachmann, Alexander Krull, Sebastian Nowozin, Jamie Shotton, Frank
  Michel, Stefan Gumhold, and Carsten Rother.
\newblock {DSAC} - differentiable {RANSAC} for camera localization.
\newblock In {\em CVPR}, 2017.

\bibitem{Brachmann2019b}
Eric Brachmann and Carsten Rother.
\newblock Neural-guided {RANSAC}: Learning where to sample model hypotheses.
\newblock In {\em ICCV}, 2019.

\bibitem{Chin2017}
Tat{-}Jun Chin, Pulak Purkait, Anders~P. Eriksson, and David Suter.
\newblock Efficient globally optimal consensus maximisation with tree search.
\newblock {\em {IEEE} Transactions on Pattern Anaysis and Machine
  Intelligence}, 2017.

\bibitem{Choi2015}
Sungjoon Choi, Qian-Yi Zhou, and Vladlen Koltun.
\newblock Robust reconstruction of indoor scenes.
\newblock In {\em CVPR}, 2015.

\bibitem{choy20194d}
Christopher Choy, JunYoung Gwak, and Silvio Savarese.
\newblock {4D} spatio-temporal {ConvNets}: Minkowski convolutional neural
  networks.
\newblock In {\em CVPR}, 2019.

\bibitem{openucn}
Christopher Choy and Junha Lee.
\newblock Open universal correspondence network.
\newblock \url{https://github.com/chrischoy/open-ucn}, 2019.

\bibitem{Choy2019}
Christopher Choy, Jaesik Park, and Vladlen Koltun.
\newblock Fully convolutional geometric features.
\newblock In {\em ICCV}, 2019.

\bibitem{UCN2016}
Christopher~B Choy, JunYoung Gwak, Silvio Savarese, and Manmohan Chandraker.
\newblock Universal correspondence network.
\newblock In {\em Advances in Neural Information Processing Systems}, 2016.

\bibitem{Chum2005}
Ondrej Chum and Jiri Matas.
\newblock Matching with prosac-progressive sample consensus.
\newblock In {\em CVPR}, 2005.

\bibitem{Dai2017}
Angela Dai, Matthias Nie{\ss}ner, Michael Zoll{\"o}fer, Shahram Izadi, and
  Christian Theobalt.
\newblock {B}undle{F}usion: Real-time globally consistent 3d reconstruction
  using on-the-fly surface re-integration.
\newblock {\em ACM Transactions on Graphics}, 2017.

\bibitem{Dang2018}
Zheng Dang, Kwang~Moo Yi, Yinlin Hu, Fei Wang, Pascal Fua, and Mathieu
  Salzmann.
\newblock Eigendecomposition-free training of deep networks with zero
  eigenvalue-based losses.
\newblock In {\em ECCV}, 2018.

\bibitem{Dong2019}
Wei Dong, Jaesik Park, Yi Yang, and Michael Kaess.
\newblock {GPU} accelerated robust scene reconstruction.
\newblock In {\em IROS}, 2019.

\bibitem{Fischler1981}
Martin~A. Fischler and Robert~C. Bolles.
\newblock Random sample consensus: A paradigm for model fitting with
  applications to image analysis and automated cartography.
\newblock {\em Communications of the ACM}, 1981.

\bibitem{Fitzgibbon2003}
Andrew~W. Fitzgibbon.
\newblock Robust registration of {2D} and {3D} point sets.
\newblock {\em Image and Vision Computing}, 2003.

\bibitem{Glocker2013}
Ben Glocker, Shahram Izadi, Jamie Shotton, and Antonio Criminisi.
\newblock Real-time {RGB-D} camera relocalization.
\newblock In {\em ISMAR}, 2013.

\bibitem{he2016deep}
Kaiming He, Xiangyu Zhang, Shaoqing Ren, and Jian Sun.
\newblock Deep residual learning for image recognition.
\newblock In {\em CVPR}, 2016.

\bibitem{heinly2015reconstructing}
Jared Heinly, Johannes~L Schonberger, Enrique Dunn, and Jan-Michael Frahm.
\newblock Reconstructing the world in six days.
\newblock In {\em CVPR}, 2015.

\bibitem{Horowitz2014}
M.~B. Horowitz, N. Matni, and J.~W. Burdick.
\newblock Convex relaxations of {SE(2)} and {SE(3)} for visual pose estimation.
\newblock In {\em ICRA}, 2014.

\bibitem{Hoseinnezhad2011}
Reza Hoseinnezhad and Alireza Bab-Hadiashar.
\newblock An m-estimator for high breakdown robust estimation in computer
  vision.
\newblock {\em Computer Vision and Image Understanding}, 2011.

\bibitem{Huber2011}
Peter~J Huber.
\newblock {\em Robust statistics}.
\newblock Springer, 2011.

\bibitem{batchnorm}
Sergey Ioffe and Christian Szegedy.
\newblock Batch normalization: Accelerating deep network training by reducing
  internal covariate shift.
\newblock In {\em ICML}, 2015.

\bibitem{Izatt2017}
Gregory Izatt and Russ Tedrake.
\newblock Globally optimal object pose estimation in point clouds with
  mixed-integer programming.
\newblock In {\em International Symposium on Robotics Research}, 2017.

\bibitem{Lebeda2012}
Karel Lebeda, Jir{\i} Matas, and Ondrej Chum.
\newblock Fixing the locally optimized {RANSAC}--full experimental evaluation.
\newblock In {\em British Machine Vision Conference}, 2012.

\bibitem{Li2009}
Hongdong Li.
\newblock Consensus set maximization with guaranteed global optimality for
  robust geometry estimation.
\newblock In {\em ICCV}, 2009.

\bibitem{Lowe2004SIFT}
David~G. Lowe.
\newblock Distinctive image features from scale-invariant keypoints.
\newblock {\em International Journal of Computer Vision}, 2004.

\bibitem{Maron2016}
Haggai Maron, Nadav Dym, Itay Kezurer, Shahar Kovalsky, and Yaron Lipman.
\newblock Point registration via efficient convex relaxation.
\newblock {\em ACM Transactions on Graphics}, 2016.

\bibitem{openmp08}
{OpenMP Architecture Review Board}.
\newblock {OpenMP} application program interface version 3.0, May 2008.

\bibitem{Pais2019}
G.~Dias Pais, Pedro Miraldo, Srikumar Ramalingam, Venu~Madhav Govindu,
  Jacinto~C. Nascimento, and Rama Chellappa.
\newblock {3DRegNet}: {A} deep neural network for {3D} point registration.
\newblock {\em arXiv}, 2019.

\bibitem{Park2017}
Jaesik Park, Qian-Yi Zhou, and Vladlen Koltun.
\newblock Colored point cloud registration revisited.
\newblock In {\em ICCV}, 2017.

\bibitem{Probst2019}
Thomas Probst, Danda~Pani Paudel, Ajad Chhatkuli, and Luc~Van Gool.
\newblock Unsupervised learning of consensus maximization for {3D} vision
  problems.
\newblock In {\em CVPR}, 2019.

\bibitem{Qi2017}
Charles~R. Qi, Hao Su, Kaichun Mo, and Leonidas~J. Guibas.
\newblock Pointnet: Deep learning on point sets for {3D} classification and
  segmentation.
\newblock In {\em CVPR}, 2017.

\bibitem{Raguram2013}
Rahul Raguram, Ondrej Chum, Marc Pollefeys, Jiri Matas, and Jan{-}Michael
  Frahm.
\newblock {USAC}: {A} universal framework for random sample consensus.
\newblock {\em {IEEE} Transactions on Pattern Anaysis and Machine
  Intelligence}, 2013.

\bibitem{Ranftl2018}
Ren{\'{e}} Ranftl and Vladlen Koltun.
\newblock Deep fundamental matrix estimation.
\newblock In {\em ECCV}, 2018.

\bibitem{unet}
Olaf Ronneberger, Philipp Fischer, and Thomas Brox.
\newblock U-net: Convolutional networks for biomedical image segmentation.
\newblock In {\em MICCAI}, 2015.

\bibitem{Rosen2019}
David~M Rosen, Luca Carlone, Afonso~S Bandeira, and John~J Leonard.
\newblock Se-sync: A certifiably correct algorithm for synchronization over the
  special euclidean group.
\newblock {\em International Journal of Robotics Research}, 2019.

\bibitem{Rousseeuw1984}
Peter~J. Rousseeuw.
\newblock Least median of squares regression.
\newblock {\em Journal of the American Statistical Association}, 1984.

\bibitem{Rusu2009FPFH}
Radu~Bogdan Rusu, Nico Blodow, and Michael Beetz.
\newblock Fast point feature histograms ({FPFH}) for 3d registration.
\newblock In {\em ICRA}, 2009.

\bibitem{Schoenberger2016sfm}
Sch\"{o}nberger, Johannes Lutz, and Jan-Michael Frahm.
\newblock Structure-from-motion revisited.
\newblock In {\em CVPR}, 2016.

\bibitem{Tennakoon2016}
Ruwan~B. Tennakoon, Alireza Bab{-}Hadiashar, Zhenwei Cao, Reza Hoseinnezhad,
  and David Suter.
\newblock Robust model fitting using higher than minimal subset sampling.
\newblock {\em {IEEE} Transactions on Pattern Anaysis and Machine
  Intelligence}, 2016.

\bibitem{thomee2016yfcc100m}
Bart Thomee, David~A Shamma, Gerald Friedland, Benjamin Elizalde, Karl Ni,
  Douglas Poland, Damian Borth, and Li-Jia Li.
\newblock {YFCC100M}: The new data in multimedia research.
\newblock {\em Communications of the ACM}, 2016.

\bibitem{Torr2000}
Philip~HS Torr and Andrew Zisserman.
\newblock {MLESAC}: A new robust estimator with application to estimating image
  geometry.
\newblock {\em Computer Vision and Image Understanding}, 2000.

\bibitem{Torr2002}
Philip H.~S. Torr.
\newblock Bayesian model estimation and selection for epipolar geometry and
  generic manifold fitting.
\newblock {\em International Journal of Computer Vision}, 2002.

\bibitem{xiao2013sun3d}
Jianxiong Xiao, Andrew Owens, and Antonio Torralba.
\newblock {Sun3D}: A database of big spaces reconstructed using sfm and object
  labels.
\newblock In {\em ICCV}, 2013.

\bibitem{Yang2013}
Jiaolong Yang, Hongdong Li, and Yunde Jia.
\newblock {Go-ICP}: Solving {3D} registration efficiently and globally
  optimally.
\newblock In {\em ICCV}, 2013.

\bibitem{Yang2014}
Jiaolong Yang, Hongdong Li, and Yunde Jia.
\newblock Optimal essential matrix estimation via inlier-set maximization.
\newblock In {\em ECCV}, 2014.

\bibitem{Yi2018}
Kwang~Moo Yi, Eduard Trulls, Yuki Ono, Vincent Lepetit, Mathieu Salzmann, and
  Pascal Fua.
\newblock Learning to find good correspondences.
\newblock In {\em CVPR}, 2018.

\bibitem{Zaheer2017deep}
Manzil Zaheer, Satwik Kottur, Siamak Ravanbakhsh, Barnabas Poczos, Ruslan~R
  Salakhutdinov, and Alexander~J Smola.
\newblock Deep sets.
\newblock In {\em Advances in Neural Information Processing Systems}. 2017.

\bibitem{Zeng20163dmatch}
Andy Zeng, Shuran Song, Matthias Nie{\ss}ner, Matthew Fisher, Jianxiong Xiao,
  and Thomas Funkhouser.
\newblock {3DMatch}: Learning local geometric descriptors from {RGB-D}
  reconstructions.
\newblock In {\em CVPR}, 2017.

\bibitem{Zhang2019}
Jiahui Zhang, Dawei Sun, Zixin Luo, Anbang Yao, Lei Zhou, Tianwei Shen, Yurong
  Chen, Long Quan, and Hongen Liao.
\newblock Learning two-view correspondences and geometry using order-aware
  network.
\newblock In {\em ICCV}, 2019.

\bibitem{Zhang1998}
Zhengyou Zhang.
\newblock Determining the epipolar geometry and its uncertainty: {A} review.
\newblock {\em International Journal of Computer Vision}, 1998.

\bibitem{Zhou2016}
Qian-Yi Zhou, Jaesik Park, and Vladlen Koltun.
\newblock Fast global registration.
\newblock In {\em ECCV}, 2016.

\bibitem{Zhou2018}
Qian-Yi Zhou, Jaesik Park, and Vladlen Koltun.
\newblock {Open3D}: {A} modern library for {3D} data processing.
\newblock {\em arXiv}, 2018.

\end{thebibliography}
}

\end{document}